\documentclass[conference]{IEEEtran}
\IEEEoverridecommandlockouts
\usepackage{cite}
\usepackage{amsmath,amssymb,amsfonts}
\usepackage{algorithmic}
\usepackage{graphicx}
\usepackage{textcomp}

\usepackage{booktabs}
\usepackage{url}
\usepackage{tikz}
\usepackage{xspace}

\newcommand\rfig[1]{Fig.\,\ref{#1}}
\newcommand\rsec[1]{Sec.\,\ref{#1}}
\newcommand\rtab[1]{Tab.\,\ref{#1}}

\newcommand\qnn[2]{\ensuremath{W^{#1}A^{#2}}}

\makeatletter
\newsavebox{\@euflag}
\sbox{\@euflag}{\raisebox{-9.8mm}{\resizebox{!}{1.2cm}{
\begin{tikzpicture}
\fill[fill={rgb,255:red,0;green,51;blue,153}] (-27,-18) rectangle (27,18);
\pgfmathsetmacro\inr{tan(36)/cos(18)}
\foreach \i in {0,1,...,11} {
  \begin{scope}[shift={(30*\i:12)}]
    \fill[fill={rgb,255:red,255;green,204;blue,0}] (90:2)
	\foreach \x in {0,1,...,4} { -- (90+72*\x:2) -- (126+72*\x:\inr) };
  \end{scope}
}
\end{tikzpicture}
}}}
\newcommand\euflag{\usebox{\@euflag}\xspace}
\makeatother

\begin{document}
\pdfinfo{
  /Title    (Inference of Quantized Neural Networks on Heterogeneous All-Programmable Devices)
  /Author   (Thomas B. Preusser, Giulio Gambardella, Nicholas Fraser, Michaela Blott)
  /Subject  (Next-Generation Processors and Architectures for Deep Learning)
  /Keywords (all-programmable, quantized neural networks, object detection)
}

\title{%
  Inference of Quantized Neural Networks\\
  on Heterogeneous All-Programmable Devices
}

\author{%
  \IEEEauthorblockN{Thomas B. Preu{\ss}er}
  \IEEEauthorblockA{%
    {\footnotesize Marie Sk{\l}odowska-Curie Fellow}\\
    \textit{Xilinx Research Labs}\\
    Dublin, Ireland\\
    thomas.preusser@utexas.edu
  }
  \thanks{\hspace*{-\parindent}
    \euflag\parbox[t]{.78\linewidth}{%
      This project has received funding from the European Union's Framework
      Programme for Research and Innovation Horizon 2020 (2014-2020) under
      the Marie Sk{\l}odowska-Curie Grant Agreement No.\,751339.
    }
  }
  \and
  \IEEEauthorblockN{Giulio Gambardella}
  \IEEEauthorblockA{%
    \textit{Xilinx Research Labs}\\
    Dublin, Ireland\\
    giulio.gambardella@xilinx.com
  }
  \and
  \IEEEauthorblockN{Nicholas Fraser}
  \IEEEauthorblockA{%
    \textit{Xilinx Research Labs}\\
    Dublin, Ireland\\
    nicholas.fraser@xilinx.com
  }
  \and
  \IEEEauthorblockN{Michaela Blott}
  \IEEEauthorblockA{%
    \textit{Xilinx Research Labs}\\
    Dublin, Ireland\\
    michaela.blott@xilinx.com
  }
}

\maketitle

\begin{abstract}
  Neural networks have established as a generic and powerful means to
  approach challenging problems such as image classification, object
  detection or decision making. Their successful employment foots on
  an enormous demand of compute. The quantization of network
  parameters and the processed data has proven a valuable measure to
  reduce the challenges of network inference so effectively that the
  feasible scope of applications is expanded even into the embedded
  domain.

  This paper describes the making of a real-time object detection in a
  live video stream processed on an embedded all-programmable
  device. The presented case illustrates how the required processing
  is tamed and parallelized across both the CPU cores and the
  programmable logic and how the most suitable resources and powerful
  extensions, such as NEON vectorization, are leveraged
  for the individual processing steps. The crafted result is an
  extended Darknet framework implementing a fully integrated,
  end-to-end solution from video capture over object
  annotation to video output applying neural network inference at
  different quantization levels running at 16~frames per second on an
  embedded Zynq UltraScale+ (XCZU3EG) platform.
\end{abstract}

\begin{IEEEkeywords}
  all-programmable, quantized neural networks, object detection
\end{IEEEkeywords}

\section{Introduction}
Neural networks have proven to be a capable and generic
machine-learning means to address
hard or even otherwise intractable problems. They have been especially
successful in image recognition, object detection and decision
making. Standard benchmarks for the evaluation of network
implementations therefore include, less surprisingly, challenges like
MNIST \cite{deng:2012} for the recognition of handwritten digits and
ImageNet \cite{deng:2009,russakovsky:2015} for the classification of
whole images. Another visual but more demanding task is the object
detection, which aims at classifying and localizing individual objects
within images. Standard reference datasets for this challenge are the
Pascal Visual Object Classes (Pascal~VOC)
\cite{everingham:2010,everingham:2015}. A scientific breakthrough and
prominent show case of competitive decision making was the AlphaGo
vs. Lee Sedol challenge match in the game of Go.

This work focuses on the object detection based
on the Pascal~VOC using a convolutional neural network (CNN) drafted
after the example of Tiny\,YOLO \cite{redmon:2016}. We focus on the
acceleration of the network inference, i.e. the employment of the
network performing its designated task, aiming at the online
processing of live video within an embedded all-programmable
platform. While our network must be trained and, indeed, re-trained to
recuperate loss of accuracy through quantization, we perform this important but
single-time effort without any exceptional resource constraints on
standard GPU hardware.

\begin{figure}
  \centerline{\begin{tikzpicture}[scale=.72,transform shape]
\draw[line width=2](-1,-1,1) rectangle (1,1,1);
\draw[line width=2](-1,1,1) -- (-1,1,-1);
\draw[line width=2](1,1,1) -- (1,1,-1);
\draw[line width=2](1,-1,1) -- (1,-1,-1);
\draw[line width=2](1,-1,-1) -- (1,1,-1);
\draw[line width=2](1,1,-1) -- (-1,1,-1);
\draw[line width=2,dashed](-1,-1,1) -- (-1,-1,-1);
\draw[line width=2,dashed](-1,1,-1) -- (-1,-1,-1);
\draw[line width=2,dashed](1,-1,-1) -- (-1,-1,-1);
\foreach \i in {-1,...,1} {
  \draw(\i,1,-1.5) -- (\i,1,1.5);
  \draw(-1.5,1,\i) -- (1.5,1,\i);
  \draw(1,\i,-1.5) -- (1,\i,1.5);
  \draw(1,-1.5,\i) -- (1,1.5,\i);

  \draw(-2,3,\i) -- (3,3,\i);
  \draw(3,-2,\i) -- (3,3,\i);
}
\foreach \i in {-2,...,3} {
  \draw(\i,-2,1) -- (\i,3,1);
  \draw(-2,\i,1) -- (3,\i,1);
  \draw(3,\i,1) -- (3,\i,-1);
  \draw(\i,3,1) -- (\i,3,-1);
}

\draw[<->](3.1,-2.1,-1) -- (3.1,-2.1,1) node[midway,below,rotate=45]{Channels $C$};
\draw[<->](-1,-2.2,1) -- (1,-2.2,1) node[midway,below]{Kernel Size $K$};
\draw[<->](-2.2,-2,1) -- (-2.2,3,1) node[midway,above,rotate=90]{Input Feature Map Size $N$};
\end{tikzpicture}}
  \caption{Feature Map Convolution}
  \label{figConvolution}
\end{figure}
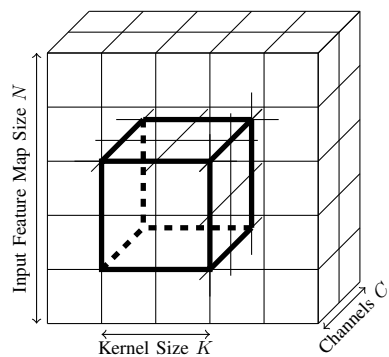
The inference performed by a CNN is computationally dominated by a
sequence of convolutions over 3-dimensional data volumes called
feature maps. A convolutional layer is often immediately followed by a
normalization, a non-linear activation and a pooling operation. While
this computation is very well structured, it is also very
intense. Particularly the convolution requires the computation of
large dot products between the network parameters, i.e. the kernel
weights, and the feature map elements. Assuming a convolutional kernel
of size $K\times K$ and a feature map depth of $C$ channels,
these dot products comprise $K^2\cdot C$ numeric multiplications for
each application of a kernel over the width and height dimensions of the
input feature map. This process is illustrated by
\rfig{figConvolution}. It is further duplicated for the same input
feature map for each of the $C'$ channels of the output feature map
using the corresponding set of kernel parameters.

A typical way to approach the convolution is its reduction to a matrix
multiplication. The rows of the multiplier matrix are constructed by
linearizing the weight parameters of the individual convolution
kernels. The number of rows equals the count of output channels to
produce. The columns of the multiplicand are correspondingly
linearized kernel application footprints so that the result matrix
will contain one convolution result in each of its elements. The
multiplicand is generated by a procedure referred to as
\texttt{im2col}. It regularly inflates the data of the input feature
map significantly. Particularly, when the kernel size is
small compared to the size of the input feature map and the stride of
its application is one, the overlap of the kernel footprints
causes \texttt{im2col} to essentially inflate the data volume by
a factor of $K^2$. On the other extreme, a convolutional kernel of the
same size of the input feature map degenerates into a single
application and, thus, a fully connected layer with no input inflation
at all.

The challenges that must be addressed by a CNN inference engine are
the storage of and timely access to the network parameters as well as
the enormous dot-product compute. Both challenges can be defused by
quantization. Eliminating unnecessary precision from the network
parameters reduces their memory footprint accordingly. Also, the
multiply-accumulate backing the dot product computation
benefits when the weights, and ideally also the feature map data, go
from floating point to fixed point arithmetic and from wider to
narrower data types. Programmable hardware as offered on
all-programmable devices is able to effectively exploit such benefits
even below an 8-bit quantization. We will refer to such aggressively
quantized CNNs as QNNs.

In the remainder of this paper, we will give a brief overview on
relevant related work with a strong emphasis on QNNs before
describing how we gradually enabled a quantized derivation of the
Tiny\,YOLO network to perform online Pascal VOC object
detection on a live video stream in a small embedded Zynq
UltraScale+ platform by leveraging the various compute capabilities of
this heterogeneous SoC. Repeatedly identifying the most severe
bottleneck, we individually describe our countermeasures and report the
achieved performance gains. \rsec{secConclusions} summarizes the
undergone development.

\section{Related Work}
The presented work is an integration effort aiming at the optimal
exploitation of a heterogeneous embedded platform. It relies on an
hardware accelerator for the inference of quantized neural networks
produced by our FINN framework \cite{umuroglu:2017}. The general idea
of aggressive quantization, going as far as the full binarization
proposed and pioneered by Hubara et\,al. \cite{hubara:2016} as well as
by Rastegari et\,al. \cite{rastegari:2016}, has adopted significant momentum
in the FPGA community. Besides FINN, also Zhao et\,al. have proposed a
binarized neural network accelerator using Vivado~HLS targeting Zynq
devices \cite{zhao:2017}.
Recently, Moss et\,al. have reported on a binary neural network
implementation \cite{moss:2017} within a hybrid data center
environment comprising a Xeon CPU and an Arria~10 FPGA solves a
similar integration task as our work. Their solution aims at saving
power in the data center by offering an alternative to GPU
accelerators. They do not at all target the ambitious resource
limitations of embedded applications.

While full binarization has been shown to work for quite a
few applications, it also fails regularly to maintain the desired
degree of accuracy. This degradation can be countered by a
slightly more moderate network quantization. The smallest possible
retreat is ternary quantization. Suggested by Li
et\,al. \cite{li:2016}, this approach has been adopted for an FPGA
implementation by Alemdar, Prost-Boucle
et\,al. \cite{alemdar:2017,prost_boucle:2017}. The use of a wider 8-bit
quantization in CNN inference can already be considered
conservative with no relevant performance degradation. It is
considered a safe enough choice to be used for ASIC
implementations of inference engines or backing matrix multiplies
as it has been done for the TPU by Google \cite{jouppi:2017}.

Our work was driven by a permanent analysis of the system performance,
the identification of the limiting bottleneck and its
mitigation. While the quantization of the network inference was a key
technique to tame both the memory requirements of network parameters
and the compute demand, we have also exploited other works in the
course of this progress. First of all, we rely on Darknet
\cite{darknet} to provide us with an open-source neural
network application environment available in customizable C code. We
have used its show case network topologies YOLO and Tiny\,YOLO
\cite{redmon:2016} as the starting point of our development making them fit
to perform the object detection in a live video stream on an embedded
all-programmable device. From the hardware point of view, we
particularly exploit the programmable fabric of a Xilinx Zynq
UltraScale+ device \cite{zynqup} and the Arm NEON technology
\cite{neon}.

\section{Building TincyYolo}
\subsection{The Challenge}
\begin{table}
  \caption{The Challenge Posed by Tiny\,YOLO versus Tincy\,YOLO}
  \label{tabTinyYolo}
  \begin{center}
  \resizebox{\linewidth}{!}{
  \begin{tabular}{ccrrc}\toprule
                      &               & Tiny\,YOLO         & Tincy\,YOLO         &\\
                      &               & \textbf{Operations}& \textbf{Operations} &\\
    \textbf{Layer \#} & \textbf{Type} & \textbf{per Frame} & \textbf{per Frame}  & \textbf{Note}\\\midrule
    1 & conv & 149520384   &  37380096   & quant. sensitive\\\cmidrule{5-5}
    2 & pool &    173056   &         -   &\\
    3 & conv & 398721024   & 797442048   &\\
    4 & pool &     43264   &     43264   &\\
    5 & conv & 398721024   & 797442048   &\\
    6 & pool &     10816   &     10816   & $>97\%$ of Compute\\
    7 & conv & 398721024   & 398721024   &\\
    8 & pool &      2704   &      2704   & Addressable by\\
    9 & conv & 398721024   & 398721024   & Offloaded\\
   10 & pool &       676   &       676   & HW QNN\\
   11 & conv & 398721024   & 398721024   & Accelerator\\
   12 & pool &       676   &       676   &\\
   13 & conv &1594884096   & 797442048   &\\
   14 & conv &3189768192   & 797442048   &\\\cmidrule{5-5}
   15 & conv &  43264000   &  21632000   & quant. sensitive\\\midrule
   $\Sigma$& &6,971,272,984&4,445,001,496&\\\bottomrule
  \end{tabular}
  }
  \end{center}
\end{table}
\begin{table*}
  \caption{Dot-Product Workloads of QNN Applications}
  \label{tabWorkloads}
  \begin{center}
  \begin{tabular}{lr@{\,}crrc}\toprule
    & \multicolumn{4}{c}{\textbf{Ops / Frame}} & \textbf{Primary Target}\\\cmidrule(lr){2-5}
    & \multicolumn{2}{c}{\textbf{Reduced}} & \textbf{8-Bit} & \textbf{Total} & \textbf{Application}\\\midrule
    \textbf{MLP-4}        &   6.0\,M &[\qnn{1}{1}]&      -- &   6.0\,M & MNIST, NIST\\
    \textbf{CNV-6}        & 115.8\,M &[\qnn{1}{1}]&  3.1\,M & 118.9\,M & CIFAR-10, Road Signs, \dots\\
    \textbf{Tincy\,YOLO}  &4385.9\,M &[\qnn{1}{3}]& 59.0\,M &4444.9\,M & Object Detection\\\bottomrule
  \end{tabular}
  \end{center}
\end{table*}
The computational challenge of object detection as posed by Tiny\,YOLO
is best appreciated by studying \rtab{tabTinyYolo}. It takes close to
7~billion floating-point operations to process a single frame. The
vast majority of these operations can clearly be attributed to the
convolutions within the hidden layers of this network. Whereas the input
and output layers of the network have proven sensitive to
quantization, the penalty paid for an aggressive quantization of the
hidden layers in terms of detection accuracy could be contained within
3\% by successful retraining. While still accounting for over 97\% of
the overall operations, these operations were simplified enormously
by using binary weights $(-1,1)$ and 3-bit feature map data. These
are ideal circumstances for a successful acceleration by programmable
hardware. For the input and output layers, this path was less
attractive as they were best left using floating point or, at most,
quantized to 8-bit fixed-point data to avoid harsh accuracy
reductions.

\begin{figure*}
  \begin{center}
    \includegraphics[scale=.33]{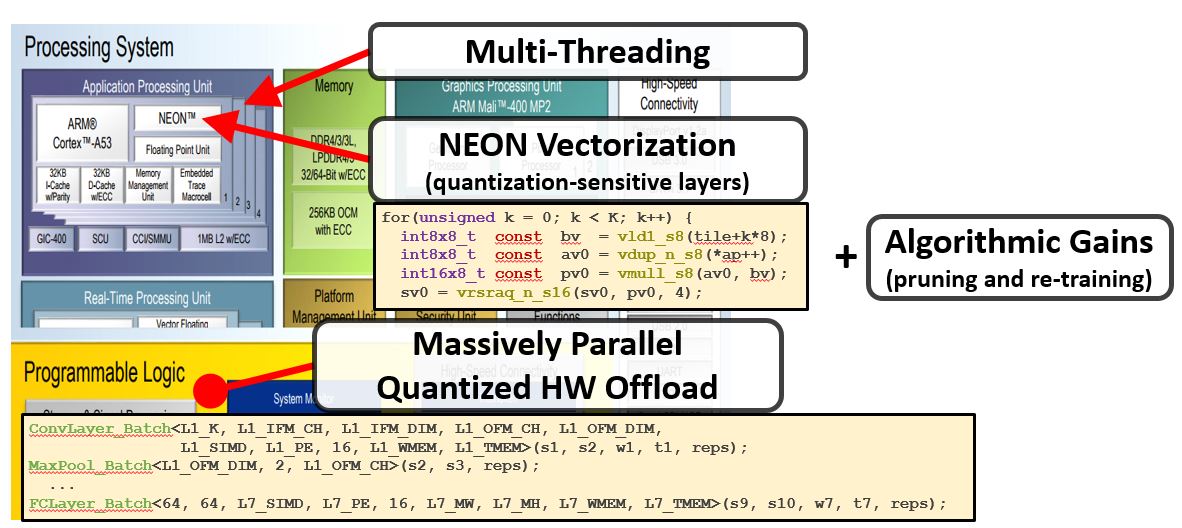}

    \footnotesize Block Diagram $[$Xilinx, Inc.$]$:
    \url{https://www.xilinx.com/content/dam/xilinx/imgs/products/zynq/zynq-eg-block.PNG}
  \end{center}
  \caption{Compute Opportunities Offered by the Zynq Platform Resources}
  \label{figPlatformOverview}
\end{figure*}
\begin{table}
  \caption{Inference Processing Time of Video Frames Broken into Stages}
  \label{tabRuntimes}
  \begin{center}\begin{tabular}{lr}\toprule
    Image Acquisition   &  40\,ms\\
    Input Layer         & 620\,ms\\
    Max Pool            & 140\,ms\\
    Hidden Layers       &9160\,ms\\
    Output Layer        &  30\,ms\\
    Box Drawing         &$\ge$15\,ms\\
    Image Output        &$\ge$25\,ms\\\midrule
    \textbf{Total}      &\textbf{10,030\,ms}\\\bottomrule
  \end{tabular}\end{center}
\end{table}
To put the computational effort in relation to previous applications
of FINN, refer to \rtab{tabWorkloads}. In anticipation of the further
operational reductions that we have performed in the process of
deriving our Tincy\,YOLO network, it already reports the reduced
operational costs of our ultimately achieved solution. These are
still greater than the previous FINN show cases by orders of magnitude
just in terms of the plain operation counts. Note that also the
individual operations are more complex as we were not able to produce
sensible results with a complete binarization of Tincy\,YOLO. While
the network weights are, indeed, binarized, we maintain a quantization
of 3~bits for all feature map values. The input layer as well as the
output layer must even process, at least, 8-bit quantities.

The significantly computational demands have a direct and severe
impact on the available implementation options. While the fully binarized
4-layer MLP and 6-layer CNN lent themselves to an implementation of
the inference engine with all layers residing one after the other in a
dataflow pipeline, this option quickly fails on resource constraints
for Tincy\,YOLO. Targeting a rather small XCZU3EG chip, only a single
generalized convolutional layer together with its subsequent
pooling layer would fit into the available fabric. The layers of the
network must be run one after the other on the same accelerator. Note
that this precludes concurrency across layers and implies a higher
latency compared to a pipeline as the feature maps between layers are
computed in full before the computation of the next layer can be
triggered.

In summary, the desired object detection poses a significantly bigger
computational challenge that is also less susceptible to
quantization.

\subsection{The Vision}
Zynq~UltraScale+ platforms are truly heterogeneous and offer more than
just an FPGA accelerator. Specifically, these platforms include a
powerful ARM multicore processor, which does not only allow to
implement applications out of a convenient OS environment but also
offers additional compute power through thread-based concurrency. In
the context of intense linear algebra, also the NEON vector extension
available in these processors is of utmost interest as it enables the
parallel SIMD operation, e.g. in four single-precision floating-point
lanes or in eight 16-bit integer lanes. These opportunities are
illustrated in \rfig{figPlatformOverview}. All of them are easily
exploitable through a C/C++-based application development. This is
even true for the hardware accelerator, which is implemented through
the HLS library of FINN. With this toolbox, we set out for taming the
embedded live object detection. Note, indeed, that the
Zynq~UltraScale+ also incorporates a Mali GPU. The exploration of its
specific workflow, toolchain and potential benefits has, however, not
been part of our work so far.

\subsection{Darknet Integration}
Building on the Tiny\,YOLO topology, using Darknet as the
training and inference framework is the natural choice. It is
open-sourced, amendable and avoids the struggle of porting and
reproducing available results within a different framework. Darknet
provides basic generic training and inference implementations in C
along with highly-optimized processing paths using CUDA-programmable
GPUs, which are its primary execution targets. Having no such GPU
available on the Zynq~UltraScale+ platform, we are starting out with
the generic inference. This delivers an disenchanting frame rate of
\textbf{0.1~fps}. Live video processing is more than two orders of
magnitude beyond reach. As shown in \rtab{tabRuntimes}, it is
the inference in the hidden network layers which contributes the
highest processing costs among all required processing stages from
image acquisition all the way to video output.

\begin{figure}
  \begin{center}
    \usetikzlibrary{shapes.geometric}
\begin{tikzpicture}[line width=1.5pt,scale=.8,transform shape]\sffamily
\coordinate(lbl) at (-6,0);
\node[draw,rectangle,line width=2pt,rounded corners,minimum width=40mm](n0) at (0,4) {\texttt{init()}};
\node[draw,rectangle,line width=2pt,rounded corners,minimum width=40mm](n1) at (0,3) {\texttt{load\_weights()}};
\node[draw,rectangle,line width=2pt,rounded corners,minimum width=40mm](n2) at (2.5,1.5) {\texttt{forward()}};
\node[draw,rectangle,line width=2pt,rounded corners,minimum width=40mm](n3) at (0,0) {\texttt{destroy()}};

\node[right,yshift=11mm] at (lbl|-n0) {Initialize Layer};
\node[right,xshift=3mm,yshift=7mm] at (lbl|-n0) {with access to};
\node[right,xshift=8mm] at (lbl|-n0) {- Configuration};
\node[right,xshift=8mm] at (lbl|-n1) {- Weight File};
\node[right] at (lbl|-n2) {Layer Inference};
\node[right] at (lbl|-n3) {Resource Cleanup};

\draw[<-] (n0) -- +(0,.8);
\draw[->] (n0) -- (n1);
\draw[->] (n1) -- (n3);
\draw[bend right=90,->] (n0|-n2) to (n2);
\draw[bend right=90] (n2) to (n0|-n2);
\draw (n3) -- +(0,-.8) node[draw,circle,fill,inner sep=2pt] (n4) {};
\end{tikzpicture}
  \end{center}
  \caption{Layer Life Cycle and Function Hooks Used by Offload Implementation}
  \label{figLayerLifeCycle}
\end{figure}

\begin{figure}
  \footnotesize
  \[\left.\begin{minipage}{.3\linewidth}
    \ttfamily
    \textbf{[convolutional]}\\
    filters=64\\
    size=3\\
    stride=1\\
    activation=relu\\
    binary=1\\

    \textbf{[maxpool]}\\
    size=2\\
    stride=2\\
    \centerline{$\vdots$}
  \end{minipage}\right\}\quad
  \begin{minipage}{.66\linewidth}
    \ttfamily
    \textbf{[offload]}\\
    \textsl{\# HW Interface Library}\\
    library=fabric.so\\
    \textsl{\# Subtopology \& Trained Weights}\\
    network=tincy-yolo-offload.json\\
    weights=binparam-tincy-yolo/\\
    \textsl{\# Output Geometry}\\
    height=13\\
    width=13\\
    channel=125
  \end{minipage}\]
  \caption{Generic Offload Mechanism Built for Darknet}
  \label{figOffload}
\end{figure}
As noted above, the inference of the hidden layers can be quantized
and assigned to a FINN-based QNN implementation. For the integration
of such an external accelerator, we implemented a generic offload
mechanism that enables Darknet to pull a particular implementation
from an arbitrary user-defined shared library.
The offload mechanism builds upon the fact that Darknet is already
virtualizing much of the layer functionality through function
pointers. Essentially, the implementation of our new offload layer
redirects those pointers to the library specified in the layer
description. Thereby the life cycle and functionality of the layer,
which is illustrated in \rfig{figLayerLifeCycle}, can be customized
completely. Note that the abstraction of such an offload layer is solely
Darknet's perspective. The backing custom implementation is only
required to compute an output feature map from a given input feature
map. Internally, it may, for instance, subsume the computation of
multiple layers of various kinds. This is practiced by our fabric
offload. The corresponding manipulation of Darknet's network
configuration is shown in \rfig{figOffload}.

Using this added offload mechanism, the QNN hardware accelerator
within the PL was integrated into the inference path of
Darknet. Although the accelerator must process one hidden layer at a
time and cannot benefit from pipelining gains due to resource
constraints, it reduces the processing time of all hidden layers
together to 30\,ms, which corresponds to a speedup of more than
$300\times$ for this particular processing stage. Taking into account
the surrounding processing, the net effect reduces to a $11\times$
speedup allowing a frame rate of just above 1~fps. It is the input
layer, which now defines the bottleneck of the computation.

\subsection{NEON Vectorization}
The generic implementation of the convolutional layers is not
optimized since Darknet targets GPU accelerators for high-performance
processing. It rather is a straightforward C implementation split into
an explicit \texttt{im2col} followed by a matrix multiplication. While
clearly being a valuable reference implementation, it must naturally
ignore platform-specific capabilities and limitations. This is the
lever available to us knowing that we target a set of ARM Cortex-A53
cores.

An obvious way to increase the number of arithmetic operations per
cycle is vectorization as offered by the NEON extension of the
platform processor. Using 128-bit registers, equivalent parallel
computations can be performed in four 32-bit lanes up to sixteen 8-bit
lanes. Also knowing that we could safely quantize the computation of
the critical first convolutional layer down to eight bits, the
employment of a NEON-optimized low-precision library appeared to be a
promising approach. Using the already developed offload mechanism, we
thus implemented a custom layer with an \texttt{im2col} implementation
that quantized the image data while arranging the multiplicand matrix
and a matrix multiplication performed through the gemmlowp library
\cite{gemmlowp}. The achieved $2.2\times$ speedup still left this
layer as the key bottleneck of the computation.

A further significant gain by optimizing the individual operations
appeared unlikely so that a fused implementation of the overall layer
was aimed at. The rationale behind this step is a significantly
increased data locality, which is especially beneficial on embedded
platforms with rather small cache sizes. So, we have sliced the
\texttt{im2col} transformation to produce the multiplicand matrix in
vertical slices. The width of these slices is matched with the number
of vector lanes that can be processed in parallel so that the
corresponding slice of the result matrix can be produced row by row
computing parallel dot products. The following input slices can
subsequently re-use the same storage over and over until the matrix
computation is complete. A generic convolutional layer implementation
following this idea achieved a $2.1\times$ speedup albeit still
operating on the original single-precision floating-point data. So,
exploiting the capabilities of NEON is itself a benefit even without
quantization.

The weight matrix of the first convolutional layer has a rather small
dimension of $16\times27$. The 16 divides nicely by all lane counts
that a NEON implementation might use, and 27 is small enough to be
unrolled explicitly. Of course, such a fully customized implementation
is no longer generic but the results are convincing. The
floating-point computation can be reduced from 620\,ms to 160\,ms, a
$3.8\times$ speedup. Re-introducing 8-bit quantization even yields
140\,ms when using a 32-bit accumulator, and 120\,ms when using a
16-bit accumulator. The 32-bit integer accumulation can actually not
utilize more vector lanes than the floating-point
implementation. However, the data locality of the 8-bit input data is
increased. The 16-bit accumulation requires a careful management of
the accumulator scale so as to avoid destructive numeric overflow in
adding up the 27~products. Therefore, a rounding right shift by 4~bit
positions must be performed before accumulation. This, in fact,
introduces some small loss of detection accuracy so that the
floating-point implementation is kept available as drop in reference
for case-to-case evaluation.

The speedup of up to $5\times$ for the convolution of the
first layer reduces the overall frame processing to
400\,ms. The implied 2.5\,fps are still not convincing. The major
bottleneck remains within the input and its subsequent maxpool layer.

\subsection{Algorithmic Simplification}
Further improvements required more daring maneuvers on the algorithmic side.
Besides the reduction of precision itself, several other changes were applied
to Tiny\,YOLO to derive Tincy\,YOLO. Specifically, the following
modifications were made: (a) leaky ReLU is replaced by ReLU;
(b) the number of output channels of layer~3 is \emph{increased} from
32 to 64; (c) the number of output channels of layers 13 \& 14 is
decreased from 1024 to 512; and (d) the first maxpool layer is removed
along with increasing the stride of the first convolutional layer from
1 to 2. The surprising but most welcome result was that after
retraining this modified network, the detection accuracy
was practically maintained. (d) alone was able to replace the two
biggest remaining bottlenecks with a lean convolution needing just
35\,ms. With this additional speedup, a frame rate of more than
5\,fps was at hand.

These algorithmic transformations are topological changes and turn
the original Tiny\,YOLO network into our Tincy\,YOLO derivative.
Accuracy scores for the modified networks are shown in Table~\ref{tab:tincy-yolo-accuracy}.
\begin{table}
  \caption{Accuracy of Tiny\,YOLO Variants}
  \label{tab:tincy-yolo-accuracy}
  \begin{center}
  \resizebox{\linewidth}{!}{
  \begin{tabular}{lcccc}\toprule
                                & Tiny         & Tiny        & Tiny         & Tincy\\
                                & YOLO         & YOLO + (a)  & YOLO + (a,b,c) & YOLO\\\midrule
    \textbf{Precision}          & Float        &[\qnn{1}{3}] & [\qnn{1}{3}] & [\qnn{1}{3}]\\
    \textbf{Accuracy mAP(\%)}   & 57.1         & 47.8        & 47.2         & 48.5\\\bottomrule
  \end{tabular}
  }
  \end{center}
\end{table}

\subsection{Parallelization}
The steps taken so far have produced a sequence of frame processing
steps that are all similarly complex. Only one of them requires the
hardware accelerator as special resource, and the most complex stage
takes 40\,ms. With a total of six stages and four available processor
cores, the theoretical maximum of a fourfold increase of the frame
rate by turning these stages into a proper processing pipeline should
only be diluted by parallelization and synchronization overhead.

Implementing the desired processing pipeline required a
complete re-implementation of Darknet's \texttt{demo} mode, which had
served well and delivered the complete end-to-end flow up to this
point. In fact, even the network inference (forward) pass had to be
disintegrated to gain access to the invocations of the individual
layers.

\begin{figure}
  \begin{center}
    \usetikzlibrary{shapes.geometric}
\begin{tikzpicture}[line width=1.5pt,scale=.8,transform shape]\sffamily
\coordinate(lbl) at (-6,0);
\node[draw,rectangle,line width=2pt,rounded corners,minimum width=40mm](n0) at (0,3.2) {Image Acquisition};
\node[draw,rectangle,line width=2pt,rounded corners,minimum width=40mm](n1) at (0,1.6) {Network Inference};
\node[draw,rectangle,line width=2pt,rounded corners,minimum width=40mm](n2) at (0,0.0) {Video Output};

\node[right] at (2,3.2) {$\left\{\begin{array}{@{\rule{4.8mm}{0mm}}c@{\rule{5.8mm}{0mm}}l}\#0&\mbox{Read Frame}\\\#1&\mbox{Letter Boxing}\end{array}\right.$};
\node[right] at (2,1.6) {$\left\{\begin{array}{cl}\#2&L[0]\\\#3&L[1]\\\vdots\\N+1&L[N-1]\end{array}\right.$};
\node[right] at (2,0.0) {$\left\{\begin{array}{cl}N+2&\mbox{Object Boxing}\\N+3&\mbox{Frame Drawing}\end{array}\right.$};

\draw[<-] (n0) -- +(0,.8) node[above] {Camera};
\draw[->] (n0) -- (n1);
\draw[->] (n1) -- (n2);
\draw[->] (n2) -- +(0,-.8) node[below] {X11};
\end{tikzpicture}
  \end{center}
  \caption{Pipeline Stages of the New \texttt{demo} Mode}
  \label{figPipeline}
\end{figure}
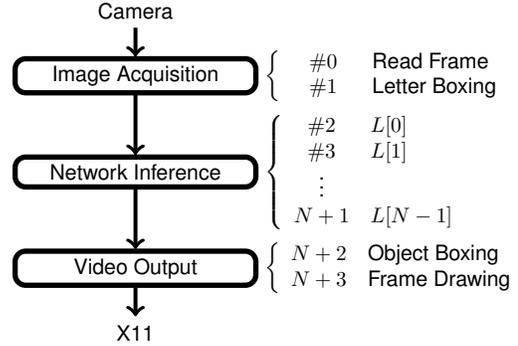
The biggest chunks of the overall computation were further split into
smaller pieces for a smoother pipeline operation. Such a move is not
sensible in a sequential frame-by-frame processing scenario as it would
only add overhead. However, in a pipelined parallel execution that
requires synchronization at the stage boundaries, the competition over
locks can be reduced beneficially by a more fine-grained division into
pipeline stages. In particular, the image acquisition was split into
the camera access and the internal scaling of the captured frame. As
illustrated in \rfig{figPipeline}, the new \texttt{demo} mode
derivative, thus, implements a pipeline that is four stages longer
than the user-specified underlying network.

For our concrete Tincy\,YOLO application, also the
implementation of the hardware offload layer was stripped of all pre-
and post-processing of its input and output data, which were therefore
moved into their own custom layer abstractions. This ensures that the
blocking of the hardware is not unduly inflated to an overgrown
containing layer abstraction but is rather limited to a tight
wrapper around the accelerated computation.

\begin{figure}
  \begin{center}
    \usetikzlibrary{shapes.geometric}
\begin{tikzpicture}[line width=1.2pt,scale=.76,transform shape]\sffamily
\node[draw,circle,fill=orange](n0) at (0,0) {};
\node[draw,rectangle,minimum width=6mm](n1) at (0,1) {};
\node[draw,circle](n2) at (0,3) {};
\node[draw,rectangle,minimum width=6mm](n3) at (0,4) {};
\node[draw,circle](n4) at (0,5) {};
\node[draw,rectangle,minimum width=6mm](n5) at (0,6) {};
\node[draw,circle,fill=green](n6) at (0,7) {};
\draw[<-](n0) -- (n1);
\draw[<-,dashed](n1) -- (n2);
\draw[<-](n2) -- (n3);
\draw[<-](n3) -- (n4);
\draw[<-](n4) -- (n5);
\draw[<-](n5) -- (n6);

\node[fill=gray!30,circle,minimum size=60mm](big) at (4,3) {};
\draw[line width=.8pt] (n2.north west) -- (big.north west);
\draw[line width=.8pt] (n2.south west) -- (big.south west);
\node[draw,rectangle,rounded corners,fill=green](free) at (4,5) {free};
\node[draw,rectangle,rounded corners](nfree) at (6,3) {$\overline{\mbox{free}}$};
\node[draw,rectangle,rounded corners,fill=orange](avail) at (4,1) {avail};
\node[draw,rectangle,rounded corners](navail) at (2,3) {$\overline{\mbox{avail}}$};
\draw[bend left=45,->](free) to (nfree);
\draw[bend left=45,->](nfree) to (avail);
\draw[bend left=45,->](avail) to (navail);
\draw[bend left=45,->](navail) to (free);
\node[rotate=90,above] at (4,3) {\textbf{Consumer}};
\node[rotate=90,below] at (4,3) {\textbf{Producer}};
\draw[line width=.5pt] (4,2) -- (4,4);
\node[xshift=4pt,right] at (2,4.1) {Starts};
\node[xshift=4pt,right] at (2,1.9) {Finishes};
\node[xshift=-4pt,left] at (6,4.1) {Starts};
\node[xshift=-4pt,left] at (6,1.9) {Finishes};
\end{tikzpicture}
  \end{center}
  \caption{Synchronization of Pipelined Frame Processing}
  \label{figScheduling}
\end{figure}
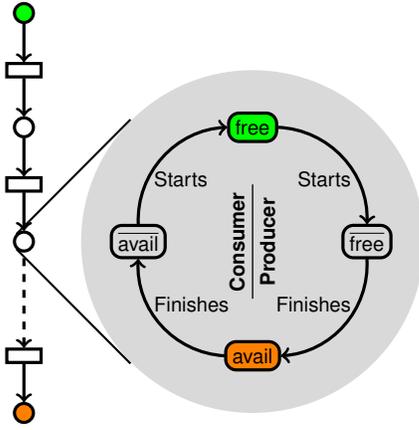
The actual processing within the pipeline is performed by a pool of
worker threads. One worker thread is allocated for each available core
and tied to it. The pipeline breaks the overall computation in
individual jobs, each of which advances the processed frame one step
further. The worker threads process one such transaction at a
time. If their current job is completed, the computed frame stays
pending in the output buffer of the corresponding pipeline stage. A
new job is selected for execution by finding the most mature one whose
output buffer is free and whose input buffer has data pending. The video
source and sink are always available and free, respectively. Note that
this scheme of job scheduling prevents that one frame overtakes
another so that the correct video sequence is maintained throughout the
processing pipeline.

The re-implemented pipelined video processing \texttt{demo} mode
achieved almost a threefold speedup resulting in a frame rate of
16\,fps. This actually allows to play live video in a way that it is
practically perceived as smooth.

\section{Conclusions}\label{secConclusions}
This paper has demonstrated how the individual heterogeneous compute
resources of a modern Zynq~Ultrascale+ platform can be systematically
exploited for implementing an Pascal VOC object detection in a live
video stream on an embedded platform. The presented measures can serve
as a blueprint to enable other machine learning applications in
resource-constrained environments. Key measures were the exploitation
of quantization, hardware acceleration, NEON vectorization,
algorithmic simplification and multi-threading for an overall speedup
of $160\times$.

\vfill
\bibliographystyle{IEEEtran}
\bibliography{IEEEabrv,literature}

\vspace*{-3.6pt}
\end{document}